\documentclass[10pt,twocolumn,letterpaper]{article}
\usepackage[pagenumbers]{wacv} % To force page numbers, e.g. for an arXiv version

\usepackage[margin=1in,columnsep=0.25in]{geometry}
\usepackage{amsmath}
\usepackage{amssymb}
\usepackage{graphicx}
\usepackage{booktabs}
\usepackage{algorithm}
\usepackage{algpseudocode}
\usepackage{xcolor}
\usepackage{caption}
\usepackage{cuted}
\usepackage[numbers,sort&compress]{natbib}

\newcommand{\net}[1]{v_\theta\!\left(#1\right)}          % gradient forward
\newcommand{\netsg}[1]{v_{\bar\theta}\!\left(#1\right)}  % stop-gradient forward
\newcommand{\shat}{\hat{s}}                               % self-conditioning signal

\definecolor{hl}{RGB}{176,58,10}

\definecolor{fmDim}{RGB}{135,142,147}
\definecolor{srcRollout}{RGB}{46,120,194}
\definecolor{srcMain}{RGB}{142,68,173}
\definecolor{srcCorrect}{RGB}{192,57,43}
\newcommand{\src}{SRC}
\newcommand{\srckc}{SRC{+}K\textnormal{-}cond}

\graphicspath{{figures/}}

\definecolor{wacvblue}{rgb}{0.21,0.49,0.74}
\usepackage[pagebackref,breaklinks,colorlinks,allcolors=wacvblue]{hyperref}

\def\wacvPaperID{40} % *** Enter the WACV Paper ID here
\def\confName{WACV}
\def\confYear{2027}

\title{Scout-Refine-Correct: Teaching Flow Models to Refine Their Own Samples}

\author{
Sergey Karpukhin\textsuperscript{1,2}
\quad\quad
Aibek Alanov\textsuperscript{1,2}\\
{\tt\small seriy.karp2@gmail.com
\quad\quad
aalanov@hse.ru}\\[0.5em]
\textsuperscript{1} FusionBrain, AXXX\\
\textsuperscript{2} HSE Unifersity\\
}

\begin{document}
\maketitle
% \begin{figure*}[!htb]
%   \centering
%   \includegraphics[width=1.1\linewidth]{figures/final_algorithm_conceptual_scheme.pdf}
%   \caption{Visual scheme of core components of SRC training algorithm.}
%   \label{fig:training-scheme}
% \end{figure*}

\begin{strip}
    \vspace{-3em}
    \centering
    \includegraphics[width=1.02\linewidth]{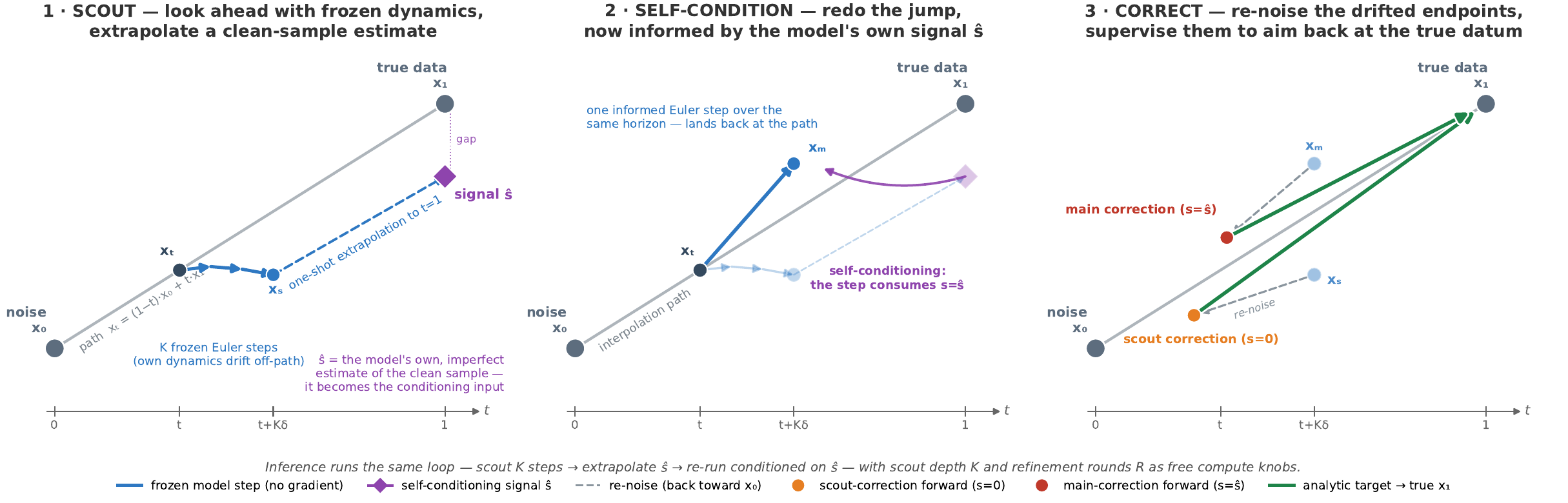}
    \captionof{figure}{Conceptual scheme of the \src{} pipeline. A scout rollout looks ahead to form a more informed estimate $\hat{x}_1$, and the main rollout then advances the current state under that condition. Correction losses train both rollouts.}
    \label{fig:algorithm_scheme}
  \end{strip}
  
  \begin{abstract}
  The quality of image generation models trained with diffusion or flow
  matching is largely fixed by the training budget: at inference, the only
  available lever is the number of sampling steps, and additional steps
  improve quality only up to a point. We present a training recipe,
  \textbf{Scout-Refine-Correct (\src{})}, that instead gives a flow-matching
  generator a genuine inference-time refinement capability. At each sampling
  step the model takes a \textbf{scout} rollout -- $K$ steps ahead along its
  own trajectory -- extrapolates a full-sample estimate, and feeds that
  estimate back to itself as a \textbf{self-conditioning} signal before taking
  the real step. This can be repeated for $R$ refinement rounds. Two
  correction losses, applied during training at states the model's own
  rollouts actually reach rather than only on the ground-truth path, teach
  the model to use this signal to reach new samples instead of collapsing to
  a suboptimal one. On ImageNet-256, \src{} improves LightningDiT-B from
  FID~\textbf{14.5} to FID~\textbf{7.26} at NFE~550, matching a
  privileged best-of-$N$ baseline that uses an \emph{oracle} FID verifier to
  choose among samples. \src{} uses no external verifier, no reward signal,
  and, after ablation, almost no extra parameters over the base architecture.
  Combining \src{} with best-of-$N$ improves the result further, to
  FID~\textbf{3.81} at NFE~4400, showing that the two axes are complementary.
  We ablate the main design choices -- the architectural pathway for the
  scout signal, the scout horizon, each correction loss, and the training
  schedule to demonstrate key working components of training.
  \end{abstract}
      
\section{Introduction}
\label{sec:intro}

A recent finding in language modeling is that accuracy can scale
with how much the model is allowed to \emph{compute at inference time} -
not just with parameter count or training compute. Chain-of-thought
prompting~\cite{wei2022chain} and self-consistency~\cite{wang2022self} spend
that compute as extra generated tokens. Later models learn the behavior
itself, either by outcome-reward reinforcement
learning~\cite{guo2025deepseek} or by supervised training on reasoning
traces~\cite{muennighoff2025s1}. No hand-written chain of thought is required
at test time: the model learns, from training, that spending more of its own
intermediate computation pays off. We use the word \emph{reasoning} in
exactly this sense throughout this paper - not textual chain-of-thought, but
\textbf{test-time compute that a trained model has learned to spend
productively on improving its own answer} - and ask whether an analogous
capability can be trained into image generators.

Flow matching~\cite{lipman2022flow,liu2022flow} is now a standard way to
train those generators, including large text-to-image
systems~\cite{esser2024scaling,labs2025flux}, and it extends to
video~\cite{jin2025pyramidal} and to discrete
data~\cite{gat2024discrete,lee2026flow}. Self-conditioning~\cite{analogbits},
in which the model conditions on its own previous estimate, is a natural
place to spend extra computation inside the sampler. Prior attempts to bring
reasoning to image generation usually rely on working with prompt conditioning. Reflection and prompt-refinement
pipelines~\cite{li2025reflect,kou2026think} iterate with an auxiliary model,
while the image model's own forward computation is untouched. Closer to our
setting, inference-time scaling has been studied by sampling many candidates
and selecting among them with a verifier~\cite{ma2025inference} - this is
also the source of the \textbf{FID-oracle best-of-$N$} baseline we compare
against throughout this paper. Critically, that approach is a \emph{pure
inference-time} search: no training changes, and it is bounded by whatever
diversity the base model's sampling distribution already contains. It
therefore cannot discover behavior the base model was never trained to
produce. Looped and recursive networks build the extra computation into the
architecture instead~\cite{dehghani2018universal,fan2025looped,geiping2026scaling,jolicoeur2025less,chng2026looped}.

Our direction is to push the refinement signal into the \emph{training}
objective itself, for \textbf{pure class-conditional} image generation
(deliberately simpler than text-to-image, so the effect is not confounded
with prompt-level search). We keep the architecture close to the base model
and implement the loop at the level of trajectory integration. We train the
model so that, at inference, it can look $K$ steps ahead along its own
sampling trajectory, extrapolate a tentative full-sample estimate from that
look-ahead, and condition its very next step on that estimate - with two
correction losses supervising exactly the states this look-ahead-and-condition
loop actually visits, while not relying on external reward model, human feedback or architectural modifications to the model.

\subsection{Contributions}
\label{sec:contrib}

\begin{itemize}
  \item A training-and-inference recipe that gives flow-matching image generators a
  genuine inference-time refinement capability, requiring no reward model
  and almost no architectural overhead beyond the base model.
  \item Proposed approach demonstrates FID on ImageNet that significantly outperforms model baselines and matches an
  oracle-verifier best-of-$N$ baseline with zero verifier calls, and
  composing with best-of-$N$ for a further large improvement to a new best
  FID at matched compute.
  \item A systematic ablation of every structural design choice - the
  architectural pathway for the scout signal, the scout horizon, each
  correction/rollout loss term, and the training-stage schedule - isolating
  exactly which pieces actually contribute to model quality.
\end{itemize}

\section{Related Works}
\label{sec:related}

% [STUB -- deferred per author instruction; prose + citations to be added
%  directly by the author. Background pointers only, not rendered:
%  - 3.1 Flow Matching: Lipman et al. flow matching / rectified flow,
%    v-vs-x1 parametrization.
%  - 3.2 Reasoning Capabilities: DeepSeek-R1, CoT, self-consistency /
%    majority voting, the two prompt-refinement-as-reasoning T2I papers,
%    the Google inference-scaling-for-image-generation paper. See
%    research_notes/papers/{vamos_analysis,flow_selfrefine_analysis,
%    fixed_point_flows_analysis,2510.20818}.md for written background.
%  - 3.3 Looped Models: TRM/HRM-style recurrent reasoning architectures,
%    Looped Flows.]

\subsection{Flow Matching}
\label{sec:fm}

Flow matching \citep{lipman2022flow, liu2022flow} has emerged as a dominant paradigm for training large-scale image generators, such as StableDiffusion 3 \cite{esser2024scaling} and Flux \cite{labs2025flux}. It has been widely adopted in other domains, including video generation \citep{jin2025pyramidal} and discrete data such as language \citep{gat2024discrete, lee2026flow}. In the discrete domain one common technique that improves inference quality is self-conditioning \citep{analogbits}, which lets the model condition on its own previous output. 

\subsection{Reasoning Capabilities}
\label{sec:reason-lm}

Reasoning as a property of deep learning models originated in large language models with manual techniques like CoT \citep{wei2022chain}, self-consitency \citep{wang2022self}. Later reasoning models appeared, which embed reasoning capabilities models through RL \citep{guo2025deepseek} or SFT on reasoning traces \citep{muennighoff2025s1}. In this work, we understand reasoning as the ability of a model to allocate more computation at inference time in order to improve the task at hand. For images and diffusion models, it has first been pursued through verifier-based search \citep{ma2025inference} and through pipelines that call external models \citep{li2025reflect, kou2026think} to improve prompt alignment or instruction following in image editing. 

\subsection{Looped Models}
\label{sec:looped-models}

Looped models are a related line of recurrent architectures. Common lines of work include looped transformers \citep{dehghani2018universal, fan2025looped, geiping2026scaling}, as well as recursive models of small size trained on specific tasks \citep{jolicoeur2025less}. Recent work \cite{chng2026looped} shows that such models can be applied to text-conditioned image generation.  This models explicitly explore idea of "refining"  output through architecture network. In contrast, in this work we keep architecture almost identical to the original DiT, while loops are implemented at trajectory integration level.
\section{Method}
\label{sec:method}

We learn a conditional velocity field for flow generation on the
linear interpolation path:
\begin{equation}
\begin{split}
   & x_t = (1-t)\,x_0 + t\,x_1, \qquad \\
  & x_0 \sim \mathcal{N}(0, I), \ \ \\ 
  & x_1 \sim p_{\mathrm{data}}, \ \ t \in [0,1],
  \label{eq:path}
\end{split}
\end{equation}
whose conditional flow-matching target is $u_t = x_1 - x_0$.

\subsection{Self-Conditioning}
\label{sec:selfcond}

The first modification is self-conditioning~\cite{analogbits}. It was
originally proposed for binary and discrete diffusion, where the denoiser
receives a second input $s$: its own previous estimate of the clean sample,
alongside the usual $(x_t, t)$. Under our parametrization, the network
predicts a velocity $v_\theta(x_t, t;\, s, y)$, and the sample estimate
$s$ is obtained by extrapolating that velocity to $t{=}1$:
\begin{equation}
  \hat{s} \;=\; x_t + (1-t)\, v_\theta(x_t, t;\, s, y).
  \label{eq:extrap}
\end{equation}
During training, half of each training batch sees $s{=}0$
and the other half sees a stop-gradient one-shot estimate
$\shat$ from Eq.~\eqref{eq:extrap} with $s{=}0$. This gives the model one
free look at its unconditioned prediction and then trains it to improve on
that guess. At inference, the signal $s$ is carried and updated across the
denoising steps.

\subsubsection{In-place refine (R-scaling)}
\label{sec:in-place-refine}

\begin{figure}[!htb]
  \centering
  \includegraphics[width=\linewidth]{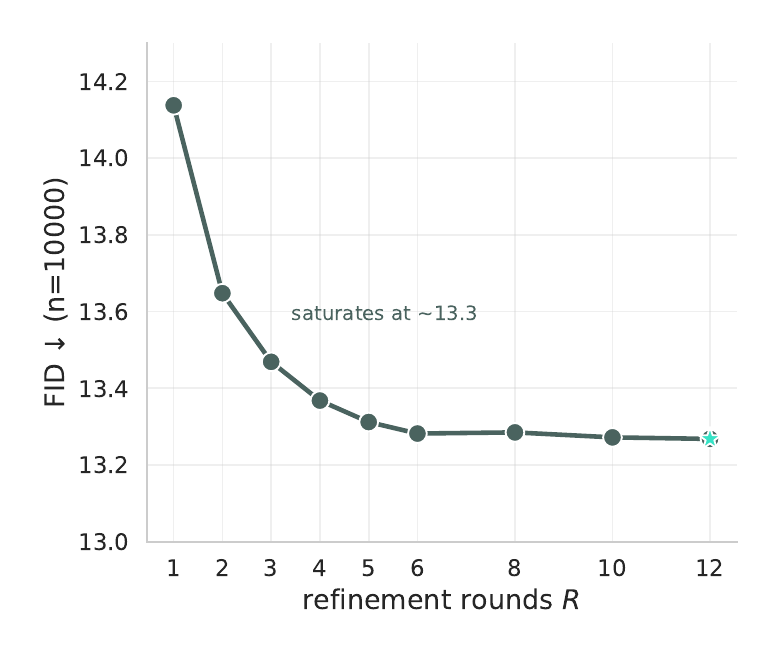}
  \caption{In-place refinement of a basic self-conditioning model (no scout,
  no correction loss): FID vs.\ refinement rounds $R$. Quality saturates at
  $\approx$13.3 almost immediately -- extra refinement compute buys nothing
  without the correction losses of Sec.~\ref{sec:self-correction}.}
  \label{fig:inplace-plateau}
\end{figure}

A natural next step is to keep iterating on $s$ at a fixed time $t$. Once
$\shat$ exists, it can be refined in place: $\shat$ is re-estimated at the
\emph{same} $(x_t,t)$ for $R$ rounds before the real step is taken. This is
fixed-point iteration on the denoiser. Similar behavior is studied
by \cite{flowreasoner}, where the desired outcome is convergence to a stable
solution $s^\star$. We do observe this convergence, but quality plateaus at
FID~$\approx$13.3 no matter how many rounds $R$ are used
(Fig.~\ref{fig:inplace-plateau}).

The reason is that training never teaches the model what to do with a
possibly wrong, self-generated $\shat$. The plain flow-matching loss is
always evaluated on the true trajectory $x_t$, so there is \textbf{no
gradient that corrects errors introduced by conditioning the model} on its
own off-trajectory rollout. This motivates the correction loss introduced
next.

\subsection{Self-Correction}
\label{sec:self-correction}

To move past the in-place refinement plateau, the loss must supervise the
model at states reached by its own, possibly imperfect rollout. These states
lie off the ground-truth interpolation path, which plain flow matching never
visits. Rather than introducing a reinforcement-learning objective with an
oracle FID reward, we follow \cite{soar} and use the following target. Take
a state produced by the model's own rollout, re-noise it toward the
\emph{same} noise $x_0$ with a random mixing factor $u\in[0,1]$, and
supervise the model at that re-noised point with the \textit{analytic}
straight-line velocity toward the true $x_1$. This pulls an off-path point
back toward the valid sample. We call the procedure \texttt{ReNoiseCorrect}
(Alg.~\ref{alg:renoise}). The main difference from SOAR is that we do not
apply classifier-free guidance when forming $x_{\mathrm{roll}}$, because we
found that it hurts training.

% \begin{figure*}[!htb]
%   \centering
%   \includegraphics[width=\linewidth]{self_correction_primer.pdf}
%   \caption{\textsc{ReNoiseCorrect} in isolation: one model rollout step off
%   the ground-truth path, one re-noise draw $z = u\,x_{\mathrm{roll}} +
%   (1-u)\,x_0$ at $t' = u\,t_{\mathrm{roll}}$, the analytic target
%   $v^\star = (x_1-z)/(1-t')$, and the resulting loss.}
%   \label{fig:selfcorr-primer}
% \end{figure*}

\begin{algorithm}[!htb]
\caption{\textsc{ReNoiseCorrect} -- turn any rollout endpoint into $N$
supervised corrections with analytic targets.}
\label{alg:renoise}
\begin{algorithmic}[1]
\Require rollout point $x_{\mathrm{roll}}$ at time $t_{\mathrm{roll}}$;
         conditioning $s$; the sample's own $(x_0, x_1, y)$
\For{$n = 1, \dots, N$}
  \State $u \sim \mathrm{Uniform}(0,1)$
  \State $z \gets u\,x_{\mathrm{roll}} + (1-u)\,x_0$; \quad
         $t' \gets u\,t_{\mathrm{roll}}$
  \State $v^\star \gets (x_1 - z)\,/\,\max(1-t',\,\sigma_{\min})$
  \State $\ell_n \gets \bigl\| \net{z, t';\, s, y} - v^\star \bigr\|_2^2$
\EndFor
\State \Return $\frac{1}{N}\sum_{n=1}^N \ell_n$
\end{algorithmic}
\end{algorithm}

\subsection{Scout-Refine-Correct}
\label{sec:scout-refine-correct}

\subsubsection{K-step scout rollout}
\label{sec:k-step-rollout}

The last design choice is to form $\shat$ from a point \emph{further along}
the trajectory, rather than from the current point or the next one. In the
\textbf{scout rollout}, the model takes $K$ null-conditioned Euler steps
forward in $t$ from $(x_t, t)$ and reaches a scout state $x_s$. One model
call at $x_s$, extrapolated with Eq.~\eqref{eq:extrap}, produces the scout
estimate $\shat$. This more informed condition then guides the single main
step.

During training, the $K$-step scout rollout is stop-gradient: it does not
send a training signal directly back to the model. 

\subsubsection{Training}
\label{sec:training}

The full training step is given in Alg.~\ref{alg:step}. It builds a
$K$-step rollout and then takes one step over the same $K\delta$ horizon,
conditioned on $\hat s$. Correction losses are applied at both points and
pull both rollouts toward the ground-truth $x_1$. $K$ is sampled once per
training step from a uniform range, but the model never observes it. We
found that explicitly conditioning on scout information, such as the number
of steps $K$, does not improve the model. In addition to the two correction
losses, the standard flow-matching loss at $(x_t, t)$ is applied to the half
of the batch whose self-conditioning signal is dropped.

Table~\ref{tab:hparams} lists the recipe's hyperparameters.
Figure~\ref{fig:algorithm_scheme} shows the main components of the step: the
$K$-rollout, the main step conditioned on $\hat s$, and the random re-noising
used by \texttt{ReNoiseCorrect}.

\begin{table}[!hbt]
\centering
\small
\begin{tabular}{llll}
\toprule
symbol & meaning & value \\
\midrule
$\rho$        & self-cond fraction of batch     & $0.5$  \\
$\delta$      & scout step size (fixed)         & $0.02$ \\
$K_{\min}..K_{\max}$ & scout-count range         & $1..3$ \\
$N$           & re-noise draws per rollout point & $2$  \\
$\lambda_{\mathrm{sc}}, \lambda_{\mathrm{m}}$ & loss weights & $1,1$  \\
\bottomrule
\end{tabular}
\caption{Training recipe hyperparameters.}
\label{tab:hparams}
\end{table}

\begin{algorithm}[t]
\caption{Single \src{} training step.}
\label{alg:step}
\begin{algorithmic}[1]
\Require batch $\{(x_1,y)\}$; network $v_\theta$; see Table~\ref{tab:hparams} for full list of hyperparameters
\State $x_0\sim\mathcal{N}(0,I)$,\quad $t\sim\mathrm{LogitNormal}(0,1)$,\quad
       $x_t\gets(1-t)x_0+tx_1$
\State $K\sim\mathrm{Uniform}\{K_{\min},\dots,K_{\max}\}$
\State \textcolor{fmDim}{$\mathcal{B}=\mathcal{S}\cup\mathcal{F}$,\quad
       $|\mathcal{S}|=\lfloor\rho|\mathcal{B}|\rfloor$}
\State \textcolor{fmDim}{$\mathcal{L}_{\mathrm{FM}}\gets
       \|\net{x_t,t;\,0,y}-(x_1-x_0)\|_2^2,\quad (x_1,y)\in\mathcal{F}$}
\Statex \vspace{2pt}\textcolor{srcRollout}{\textbf{K-rollout + signal}}
\State \textcolor{srcRollout}{$(x_s,t_s)\gets
       \Call{EulerRollout}{v_{\bar\theta},x_t,t;\,K,\delta,s{=}0}$}
\State \textcolor{srcRollout}{$\shat\gets
       x_s+(1-t_s)\,\netsg{x_s,t_s;\,0,y}$}
\Statex \vspace{2pt}\textcolor{srcMain}{\textbf{Main step with $\shat$}}
\State \textcolor{srcMain}{$x_m\gets
       x_t+(t_s-t)\,\netsg{x_t,t;\,\shat,y}$}
\Statex \vspace{2pt}\textcolor{srcCorrect}{\textbf{Correction losses}}
\State \textcolor{srcCorrect}{$\mathcal{L}_{\mathrm{sc}}\gets
       \Call{ReNoiseCorrect}{x_s,t_s;\,s{=}0}$}
\State \textcolor{srcCorrect}{$\mathcal{L}_{\mathrm{m}}\gets
       \Call{ReNoiseCorrect}{x_m,t_s;\,s{=}\shat}$}
\State $\mathcal{L}\gets(1-\rho)\bar{\mathcal{L}}_{\mathrm{FM}}
       +\rho(\lambda_{\mathrm{sc}}\bar{\mathcal{L}}_{\mathrm{sc}}
       +\lambda_{\mathrm{m}}\bar{\mathcal{L}}_{\mathrm{m}})
       +\bar{\mathcal{L}}_{\cos}$
\end{algorithmic}
\end{algorithm}

% \begin{figure}[t]
%   \centering
%   \includegraphics[width=\linewidth]{training_scheme.pdf}
%   \caption{\textbf{(A)} One training step as a left-to-right dataflow: cold
%   dashed = frozen forwards, warm = the three gradient forwards, of which
%   only the main correction (deep warm) consumes $\shat$. \textbf{(B)} The
%   network: $s$ enters only as concatenated tokens from a separate, bias-free
%   embedder; one AdaLN slot driven by $c = \mathrm{MLP}_t(t) + E_y(y)$ -- no
%   pathway for $K$ anywhere (Sec.~\ref{sec:5.3.1}).}
%   \label{fig:training-scheme}
% \end{figure}

\subsubsection{Inference}
\label{sec:inference}

Plain self-conditioning carries $\shat$ from one main step to the next,
computing it once and reusing it at no extra cost. Our training instead
produces a useful estimate $\hat s$ at times closer to the data. After
comparing inference regimes, including in-place $R$ refinement (outer loop) and scout
rollouts of depth $K$ (inner loop), we use the procedure in Alg.~\ref{alg:sample} to
search over inference budgets. The budget is computed by following formula: 
\begin{equation}
  \mathrm{NFE} \;=\; T\,\bigl(R\,(K{+}1) + 1\bigr),
  \label{eq:nfe}
\end{equation}
We fix $T{=}50$ main Euler steps in every
setup. The pair $(K,R)$ can be chosen at inference time. After searching
through grid of $(K, R)$ values and discarding unstable operating points, we end up with
high-quality FiD/NFE frontier. 

% \begin{algorithm}[t]
% \caption{SRC inference - $K{\times}R$ self-guided sampling.}
% \label{alg:sample}
% \begin{algorithmic}[1]
% \Require class $y$; scout depth $K$, rounds $R$; Euler schedule
%          $0=t_0<\dots<t_T=1$; scout step size $\delta$
% \State $x\sim\mathcal{N}(0,I)$
% \For{$i=0,\dots,T-1$}
%   \State $\shat\gets\Call{ScoutSignal}{x,t_i}$
%   \State $v\gets\net{x,t_i;\,\shat,y}$
%   \State $x\gets x+(t_{i+1}{-}t_i)\,v$ \ \ \ (\(\shat\) discarded)
% \EndFor
% \State \Return $x$
% \end{algorithmic}
% \end{algorithm}

\begin{algorithm}[t]
\caption{\src{} inference step with a $K$-step scout and $R$ refinement rounds.}
\label{alg:sample}
\begin{algorithmic}[1]
\Require $y$;\quad $\{t_i\}_{i=0}^{T}$;\quad $K,R,\delta$
\State $x\sim\mathcal{N}(0,I)$
\For{$i=0,\dots,T-1$}
  \State \textcolor{srcRollout}{$\shat_0\gets 0$}
  \For{$r=1,\dots,R$}
    \State \textcolor{srcRollout}{$\shat_r\gets
           \Call{ScoutRollout}{x,t_i,y;\,K,\delta,\shat_{r-1}}$}
  \EndFor
  \State \textcolor{srcMain}{$x\gets x+(t_{i+1}-t_i)\,
         \net{x,t_i;\,\shat_R,y}$}
\EndFor
\State \Return $x$
\end{algorithmic}
\end{algorithm}

% \begin{algorithm}[t]
% \caption{\textsc{ScoutSignal} -- $R$ rounds of scout/extrapolate at a fixed
% state, each round re-conditioned on the previous round's estimate.}
% \label{alg:scout}
% \begin{algorithmic}[1]
% \Require state $x$ at time $t$; class $y$; knobs $K,R,\delta$
% \State $\shat_0\gets 0$
% \For{$r=1,\dots,R$}
%   \State $x_s\gets x$
%   \For{$j=0,\dots,K-1$}
%     \State $x_s\gets x_s+\delta\,\net{x_s,\tau_j;\,\shat_{r-1},y}$
%   \EndFor
%   \State $\shat_r\gets x_s+(1-\tau_K)\,\net{x_s,\tau_K;\,\shat_{r-1},y}$
% \EndFor
% \State \Return $\shat_R$
% \end{algorithmic}
% \end{algorithm}

\section{Results}
\label{sec:results}

\subsection{Image generation}
\label{sec:5.1}

\begin{figure}[!htb]
  \centering
  \includegraphics[width=\linewidth]{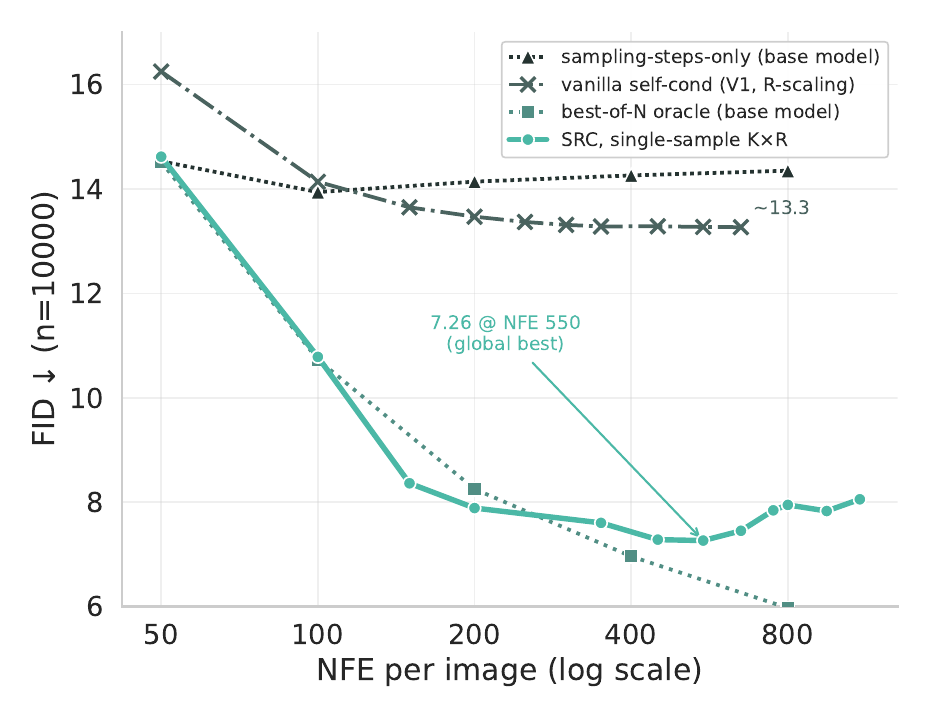}
  \caption{FID vs.\ NFE on ImageNet-256: sampling-step scaling alone (base
  model, flat), vanilla self-conditioning $R$-scaling (plateaus at
  $\approx$13.3), and \src{}'s single-sample $K{\times}R$ frontier (global
  best FID~7.26 at NFE~550).}
  \label{fig:global-comparison}
\end{figure}

We evaluate on ImageNet-256 \citep{deng2009imagenet}, and the base architecture is LightningDiT-B \citep{yao2025reconstruction}. Detailed training hyperparameters are given in
Appendix~\ref{appendix: training details}. We compare \src{} with three baselines: scaling the number
of sampling steps $T$ of the base model, scaling the in-place refinement
rounds $R$ of a self-conditioned model, and an oracle best-of-$N$ baseline
that selects the candidate which most improves the target FID. In Figure~\ref{fig:global-comparison}
it can be seen that sampling-step scaling leaves the base LightningDiT-B
checkpoint at roughly the same quality. Self-conditioning $R$-scaling
plateaus after a gain of about 3 FID. \src{} reaches a single-sample best
of FID~\textbf{7.26} at NFE~550 ($K{=}4,R{=}2$). It is competitive with the
privileged best-of-$N$ oracle even though training never observes the FID
metric. 

\textbf{Combining \src{} with best-of-$N$.} We apply best-of-$N$ to \src{}
in a refine-then-select pipeline: draw $N$ independent \src{} samples at a
fixed $(K,R)$, keep the sample selected by the FID oracle, and measure how
quality scales with $N$ at different $(K,R)$ anchors.

\begin{table}[!htb]
\centering
\small
\begin{tabular}{lccc}
\toprule
NFE & \src{} ($K{=}2,R{=}1$)+Bo$N$ & baseline Bo$N$\\
\midrule
200  & \textbf{7.89}  & 8.25 \\
400  & \textbf{6.53}  & 6.96 \\
800  & \textbf{5.45} ($N{=}4$)  & 5.97 ($N{=}8$) \\
1600 & \textbf{4.12}  & -    \\
\bottomrule
\end{tabular}
\caption{Same-budget comparison: \src{} anchored at $K{=}2,R{=}1$ versus the
baseline best-of-$N$ oracle}
\label{tab:bestofn-samebudget}
\end{table}

\emph{(a) Same-budget result.} Anchored at the inexpensive end of our
frontier ($K{=}2,R{=}1$, NFE~200), \src{} with best-of-$N$ beats the baseline
best-of-$N$ oracle at every NFE where both are measured. This includes
beating the baseline's own best-of-$N$ ceiling (FID~5.97 at $N{=}16$) with
only a quarter of that NFE and far fewer draws
(Table~\ref{tab:bestofn-samebudget}).

\emph{(b) Beyond the best-of-$N$ frontier.} Starting from the best
single-sample \src{} setting ($K{=}4,R{=}2$, NFE~550) and scaling to
$N{=}8$, \src{} with best-of-$N$ reaches FID~\textbf{3.81} at NFE~4400
(Table~\ref{tab:bestofn-beyond}). Refinement combined with verifier search
therefore both wins at a smaller NFE budget and continues to a higher-quality
operating point.

\begin{table}[t]
\centering
\small
\begin{tabular}{ccc}
\toprule
$N$ & \src{} ($K{=}4,R{=}2$)+best-of-$N$ & NFE \\
\midrule
1 & 7.26 & 550 \\
2 & 5.99 & 1100 \\
4 & 5.02 & 2200 \\
8 & \textbf{3.81} (new global best) & 4400 \\
\bottomrule
\end{tabular}
\caption{Scaling \src{}+best-of-$N$ past where the baseline best-of-$N$
oracle was ever measured.}
\label{tab:bestofn-beyond}
\end{table}

% \begin{figure*}[!htb]
%   \centering
%   \includegraphics[width=\linewidth]{bestofn_two_claims.pdf}
%   \caption{(a)~Same-budget: \src{} ($K{=}2,R{=}1$)+best-of-$N$ vs.\ the
%   baseline best-of-$N$ oracle at matched NFE. (b)~Scaling further:
%   \src{} ($K{=}4,R{=}2$)+best-of-$N$ reaches FID~3.81, past the baseline's
%   own measured range (shaded).}
%   \label{fig:bestofn-two-claims}
% \end{figure*}

\begin{figure*}[!htb]
  \centering
  \includegraphics[width=\linewidth]{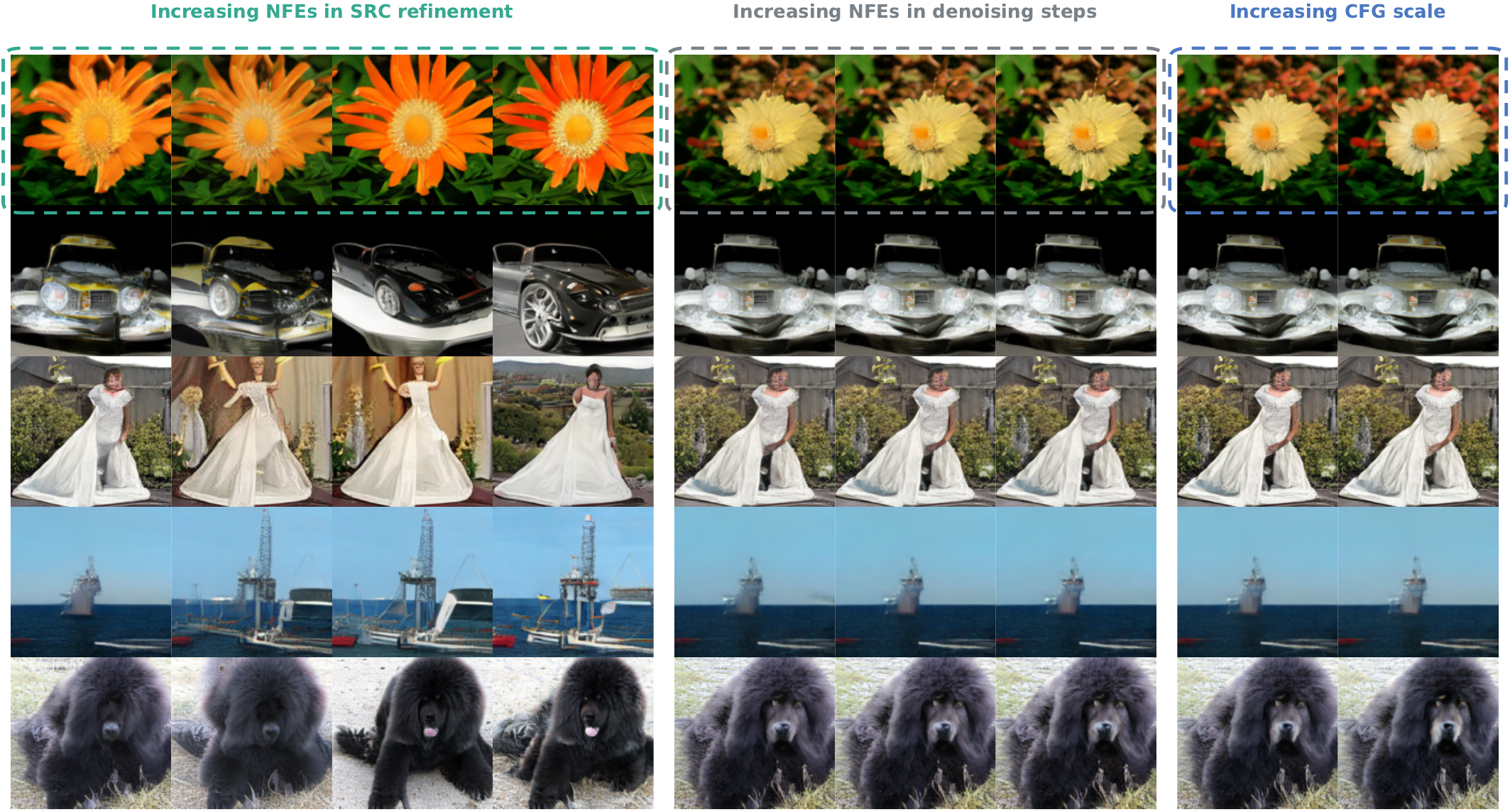}
  \caption{\src{} $K/R$ refinement compared with scaling the number of
  inference steps and with scaling the classifier-free guidance weight.}
  \label{fig:sample-grid}
\end{figure*}

\textbf{Qualitative effect on images.} Figure~\ref{fig:sample-grid} shows
that \src{} refinement visibly restructures the sampled image. Spending the
extra compute on more denoising steps, or on tuning the classifier-free
guidance scale, does not have the same effect. Figure~\ref{fig:cfg_heatmap}
checks the latter directly by showing the pixel difference after each kind
of scaling.

\subsection{Design choices}
\label{sec:design-choices}

We pose this section as a sequence of design questions around SRC training, each answered by a
targeted ablation, rather than a results dump:
\begin{enumerate}
    \item Which correction/rollout terms contribute to the resulting quality?
    \item What is the effect of dropping the self-conditioned flow-matching loss?
  \item How wide should the training-time scout horizon $K_{\max}$ be?
  \item How should self-conditioning pretraining and \src{} training be balanced?
  \item Is tuning the classifier-free-guidance scale a free substitute for
  our refinement?
\end{enumerate}

\subsubsection{Which train components contribute to the resulting quality? }

\begin{figure*}[!htb]
  \centering
  \includegraphics[width=\linewidth]{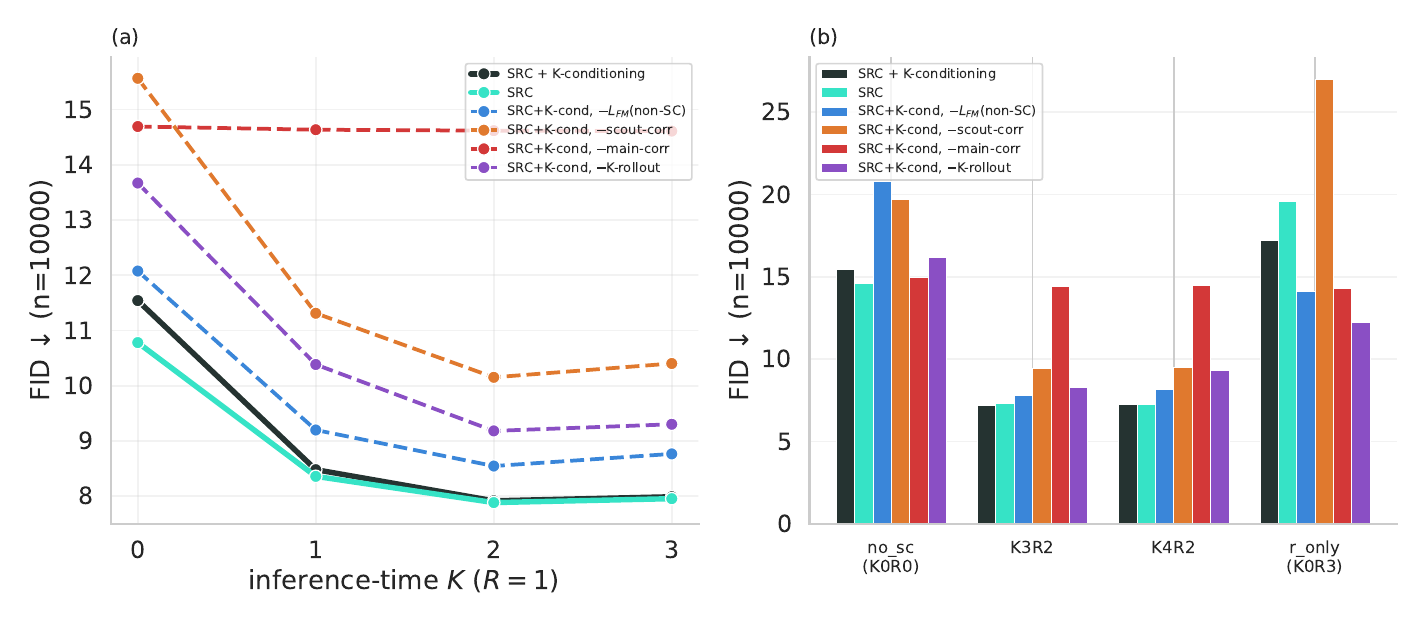}
  \caption{(a)~FID vs.\ inference-time $K$ at $R{=}1$. (b)~FID at four
  decisive operating cells. \textbf{main-corr} is the single most critical
  term; \textbf{scout-corr} is also clearly load-bearing, while $K$-conditioning doesn't introduce significant gains.}
  \label{fig:component-ablation}
\end{figure*}

In this experiment we run leave-one-out ablation of the \srckc{} model - it is slightly different from
\src{} by an explicit conditioning pathway for $K$. In Fig. \ref{fig:component-ablation} for this ablations we show scaling for fixed R, as well as some other operating points. Notably, removing conditioning on K (which is our final \src{} version) behaves almost same and can be better especially without refining. Among the remaining choices, removing \textbf{main-corr}
is by far the most damaging: FID becomes completely flat, so the model never
learns to use the refined scout condition $\hat s$ at inference time.
Removing \textbf{scout-corr} also hurts substantially, raising the best FID
to 9.43. Removing the non-self-conditioned flow-matching term or the
$K$-step rollout also degrades the quality frontier. \src{} itself appears
in the same figure as the ablation that removes the architectural change.
An additive version of this study, in which each training component is
introduced on the way from self-conditioning to \src{}, is given in
Appendix~\ref{appendix: more vis examples}.

% \begin{figure}[!htb]
%   \centering
%   \includegraphics[width=\linewidth]{nss_arch_kr_grid.pdf}
%   \caption{\src{} vs.\ \srckc{}: each curve shows the running-min achieved
%   frontier up to its own best point (no cross-family oscillation, no
%   high-$R$ blow-up tail). Removing the architectural $K$-conditioning
%   pathway entirely (\src{}) matches or beats keeping it (\srckc{}).}
%   \label{fig:nss-arch-grid}
% \end{figure}

\subsubsection{What is the effect of dropping self-conditioned flow matching loss?}
\label{fm-loss-sc}

\src{} keeps the flow-matching loss on the part of the batch that does not
use self-conditioning, but drops the flow-matching loss evaluated while the
network is conditioned on its own estimate. We compare two correction-trained
models that differ only in this term: flow matching at the current state
$(x_t, t)$ with conditioning $\shat$. The earlier variant keeps the term,
and the later drops it.

\begin{table}[!htb]
\centering
\small
\begin{tabular}{lcc}
\toprule
 & $R{=}2$ & $R{=}3$ \\
\midrule
with self-conditioned FM & 0.013 & 0.003 \\
without self-conditioned FM & 0.164 & 0.183 \\
\bottomrule
\end{tabular}
\caption{Mean LPIPS between refinement round $R$ and round $R{+}1$, averaged
over images and sampling steps. The two rows are the self-conditioning
baseline that keeps the term and the later model that drops it.}
\label{tab:fm-lpips}
\end{table}

Table \ref{tab:fm-lpips} demonstrates effect of such change, reporting LPIPS between decoded $s$ at $R$a $s$ at $R+1$. It can be observed that with the $fm$ term, that distance is already near zero by
$R{=}2$. Without it, an extra round moves the sample substantially, while the FiD rapidly improves - between $R=0$ and $R=1$ from FiD moves from 15.17 to 10.98.

\begin{figure*}[!htb]
  \centering
  \includegraphics[width=\linewidth]{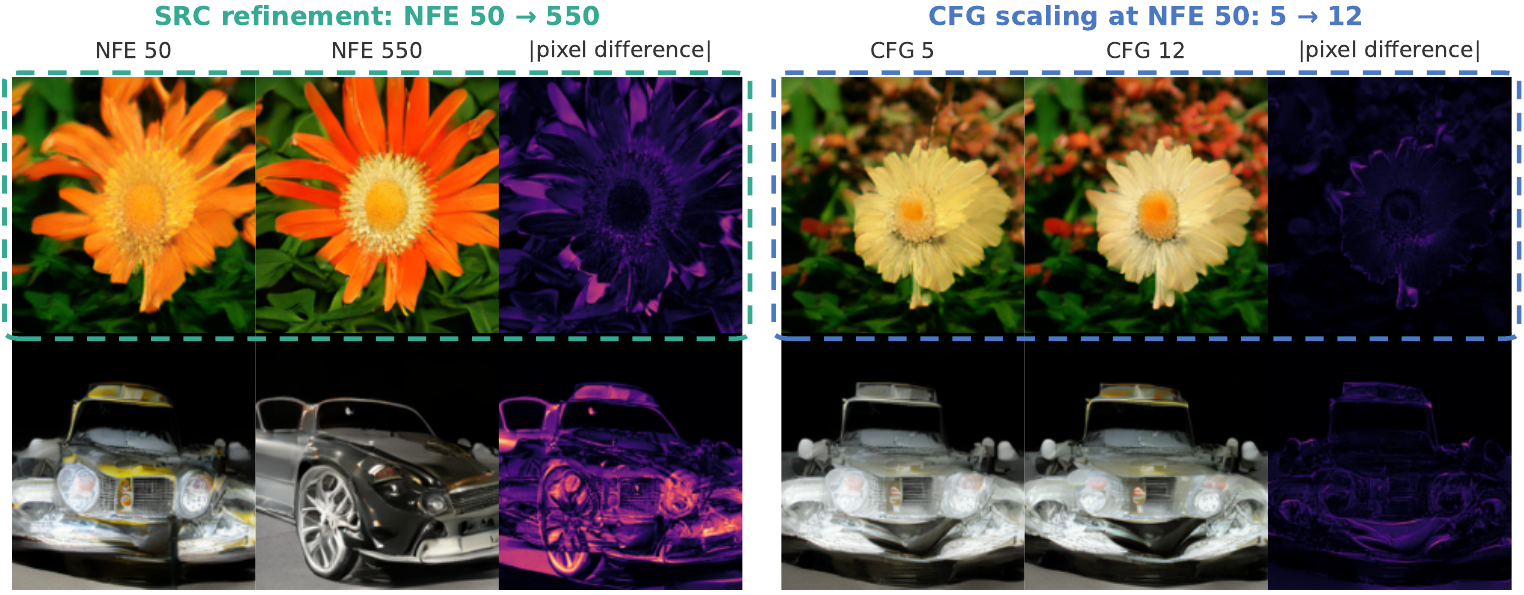}
  \caption{Pixel differences for \src{} refinement and for
  classifier-free guidance scaling, on the same examples.}
  \label{fig:cfg_heatmap}
\end{figure*}

\subsubsection{How wide should the training-time scout horizon $K_{\max}$ be?}

Table~\ref{tab:kmax} sweeps the training-time scout horizon
$K_{\max}\in\{2,3,4,6,8\}$ and reports each variant's best $(K,R)$
operating point - $K_{\max}{=}3$ is the best setting. Current interpretation of larger K not helping with scaling is that being too close to data limits exploration of the model.

\begin{table}[!htb]
\centering
\small
\begin{tabular}{cc}
\toprule
$K_{\max}$ & best FID \\
\midrule
2 & 7.21 \\
\textbf{3} & \textbf{7.18} \\
4 & 7.49 \\
6 & 8.30 \\
8 & 9.99 \\
\bottomrule
\end{tabular}
\caption{Training-time scout horizon ($K_{\max}$). $K_{\max}{=}3$ is the best of the tested values.}
\label{tab:kmax}
\end{table}

% \subsubsection{Which loss components are load-bearing?}
% \label{sec:5.3.3}

% Our first recipe fed $K$ into the network through an explicit extra pathway
% (NSS architecture: dedicated LoRA weights and an additional AdaLN slot
% conditioned on $K$) -- call this variant \srckc{}. We ablated this pathway
% entirely (\src{}, no $K$-conditioning anywhere, $K$ only ever sets the scout
% \emph{horizon} as in Algorithms~\ref{alg:step}--\ref{alg:scout}) and found,
% across the entire practical $K{\times}R$ operating basin (NFE 50--1100,
% $R\le3$), \src{} matches or beats \srckc{} -- including a new global best of
% FID~7.26 at $K{=}4,R{=}2$ (NFE~550), better than \srckc{}'s own global best
% of 7.52 at $K{=}3,R{=}2$ (NFE~450) -- using a strictly simpler architecture
% with no extra parameters (Fig.~\ref{fig:nss-arch-grid}). The one real
% weakness is $R$-scaling at high $R$ with low $K$ (the architectural pathway
% appears to buy some many-round stability), but no operating point either
% checkpoint actually uses lives past $R{=}3$, so this does not affect any
% practical deployment. We adopt \src{} (no architectural $K$-conditioning) as
% the final recipe used everywhere else in this paper, confirming that the
% training \emph{procedure} is what grants the refinement capability, not any
% added architecture.

\subsubsection{Training-stage budget ablation}
\label{sec:5.3.4}

\src{} training can be composed of following stages stages: (a) self-conditioned
pretraining, (b) \src{} with a fixed scout depth $K{=}1$, and (c) the final
scout-and-correction stage. Training runs are all limited to 210k iterations.
Table~\ref{tab:budgetsplit} compares budget splits among these stages. The
split $(120\mathrm{k},60\mathrm{k},30\mathrm{k})$ beats the four
alternatives, including both variants that skip a stage, so all three stages
contribute. 

\begin{table}[!htb]
\centering
\small
\begin{tabular}{lc}
\toprule
(a,b,c) steps & FID @ best KR \\
\midrule
(120k, 0, 90k) -- no stage b   & 7.85 \\
(120k, 90k, 0) -- no stage c   & 8.88 \\
\textbf{(120k, 60k, 30k)} & \textbf{7.18} \\
(120k, 30k, 60k)               & 7.53 \\
(90k, 90k, 30k)                & 7.34 \\
\bottomrule
\end{tabular}
\caption{Training-stage budget. The $(120\mathrm{k},60\mathrm{k},30\mathrm{k})$ split is the best of the five schedules tested.}
\label{tab:budgetsplit}
\end{table}

\subsubsection{Is cfg-scale tuning a free substitute for our refinement?}
\label{sec:5.3.5}

At a fixed budget of NFE$=$50, raising the classifier-free guidance scale improves FID from 14.52 to \textbf{9.54}. \src{} overtakes it by NFE$=$150
by reaching FID 8.36 and keeps improving while the guidance sweep
saturates. However more important distinction is qualitative - in terms of perceptual difference \src{} produces completely different images, as highlighted in
Figure~\ref{fig:cfg_heatmap} shows that \src{} refinement changes pose,
composition, and fine structure, whereas guidance-scale tuning changes only
small regions.

\section{Conclusion}
\label{sec:conclusion}

We presented Scout-Refine-Correct, a training recipe that teaches a
flow-matching image generator to spend extra computation at inference time
on its own sample. At every step the model looks $K$ steps ahead, refines
the resulting self-conditioning signal for $R$ rounds, and then takes one
informed step. Two correction losses supervise the off-path states that
these rollouts actually reach, so the additional computation changes the
sample rather than saturating.

On class-conditional ImageNet-256, \src{} improves LightningDiT-B from
FID~14 to FID~7.26 at 550 network evaluations. This matches a privileged
best-of-$N$ oracle that selects samples with the true FID statistic, while
\src{} itself uses no verifier and no reward model. Using both together
reaches FID~3.81 at 4400 evaluations. Qualitatively, the refinement changes
pose, composition, and fine detail. Increasing the number of denoising steps
or the classifier-free guidance scale does not produce the same change.

The ablations locate the source of this behavior. The main and scout
correction losses are required for the model to use the scouted signal.
Dropping flow matching on the self-conditioned state improves quality at one
refinement round, but later rounds no longer converge. The scout horizon and
the split of the three training stages both have a clear best setting, and
neither guidance-scale tuning nor a longer Euler schedule replaces the
correction training.

{
    \small
    \bibliographystyle{ieeenat_fullname}
    \bibliography{main}
}

\onecolumn
\appendix

\section{Training details}
\label{appendix: training details}

All models are trained on ImageNet-256. The path is linear flow with a
velocity target, logit-normal time sampling, and a cosine loss term. The
optimizer is AdamW with $\beta=(0.9, 0.95)$ and no weight decay. Unless
noted, evaluation uses 50 Euler steps, classifier-free guidance scale 5,
interval start 0.11, and timestep shift 0.3 (default sampling parametrs from Lightning-B)

\subsection{LightningDiT-B}
\label{app:lightningdit}

The step-scaling and best-of-$N$ baselines use LightningDiT-B/1, trained for
60k steps at batch size 1024 and learning rate $4\times 10^{-4}$. The
backbone uses rotary position embeddings, RMSNorm, and SwiGLU. It has no
self-conditioning input. Its standard 50-step sample reaches FID 14.52.

\subsection{Self-conditioning baseline}
\label{app:selfcond}

The in-place refinement baseline is trained separately for 120k steps, at
batch size 256 and learning rate $1.5\times 10^{-4}$. It is not a
fine-tune of LightningDiT-B. Relative to that backbone, rotary embeddings
are replaced by a learned positional embedding, QK-norm is turned on, and
the network gains a second input $s$, its own estimate of the clean sample.

\begin{figure*}[!htb]
  \centering
  \includegraphics[width=\linewidth]{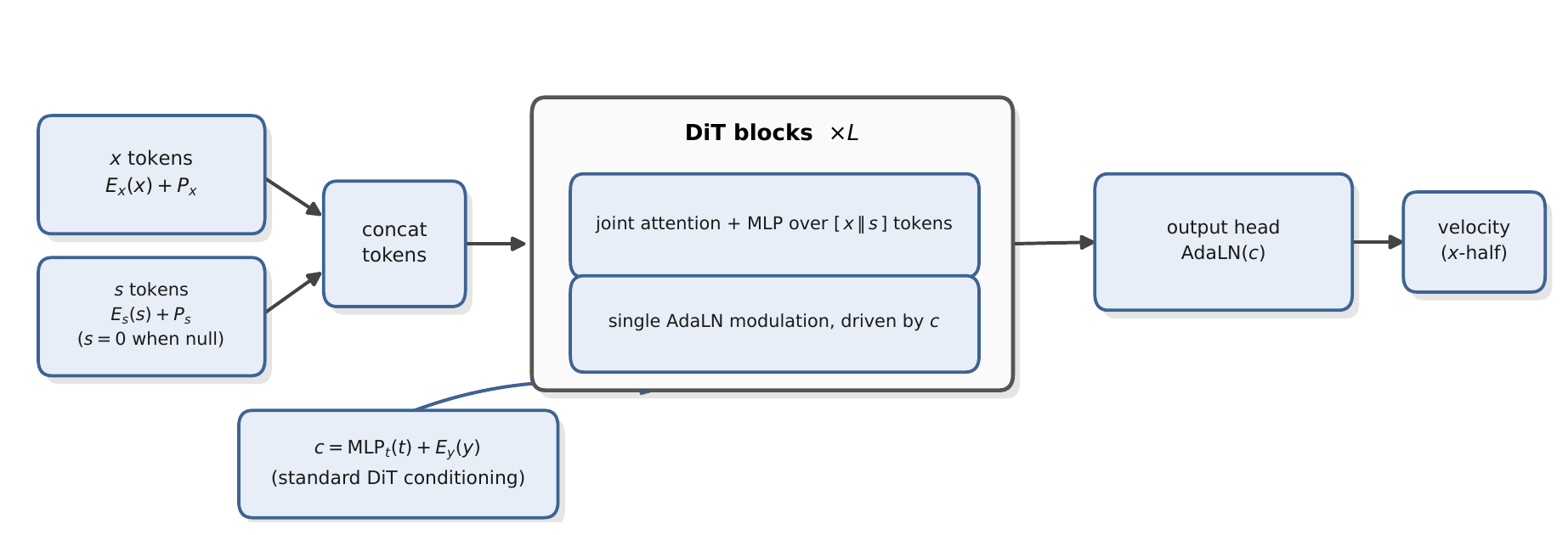}
  \caption{The architecture used from the
  self-conditioning baseline through \src{}: $s$ enters only as
  concatenated tokens from a bias-free embedder, and one AdaLN slot is
  driven by time and class. $K$ is not a network input.}
  \label{fig:arch}
\end{figure*}

Figure~\ref{fig:arch} shows where that input enters. $s$ is embedded by a
separate bias-free patch embedder, so the trained null $s{=}0$ maps to the
zero token, and those tokens are concatenated with the image tokens. A
single AdaLN slot, driven by the usual time and class vector, modulates
the blocks. There is no second AdaLN slot and no input for the scout depth
$K$. Half of each batch is trained with $s{=}0$ and half with a
stop-gradient one-step estimate, which is the standard self-conditioning
bootstrap. At inference this checkpoint carries the estimate from one step
to the next; the $R$-scaling curve refines it in place before the step.

\subsection{Three-stage \src{} schedule}
\label{app:schedule}

\src{} keeps the self-conditioning architecture and warm-starts from that
120k checkpoint. The schedule is $(120\mathrm{k}, 60\mathrm{k},
30\mathrm{k})$, 210k steps in total. Stages 2 and 3 use batch size 128 and
learning rate $1.5\times 10^{-4}$.

\textbf{Stage 2} (120k to 180k) adds the two correction losses while the
self-conditioning signal is still a same-point estimate: one frozen
prediction at the current $(x_t, t)$, with no $K$-step look-ahead. Both
the scout and the main correction are on. Each is the mean of $N{=}2$
independent re-noisings of the same rollout point. Flow matching on the
self-conditioned state is still applied, in addition to ordinary flow
matching on the half of the batch that sees $s{=}0$ (it can be turned of just fine, but keeping it for this stage was found benefitial for later). The scout step used
inside the correction is a single step of size $\delta{=}0.02$.

\textbf{Stage 3} (180k to 210k) is the final \src{} recipe. The signal is no
longer formed at the current point. $K$ is drawn once per step from
$\{1,2,3\}$, and $\hat s$ is extrapolated after $K$ frozen Euler steps of
size $0.02$. The main correction then covers that same horizon in one hop.
The self-conditioned flow-matching term is dropped: the self-conditioned
half of the batch is supervised only by the two correction losses, still
with $N{=}2$. Ordinary flow matching remains on the other half. The
architecture is unchanged, so $K$ still only sets the rollout length. All
\src{} numbers in the paper use this 210k checkpoint.

\section{More visual reasoning examples}
\label{appendix: more vis examples}

\begin{figure*}[!htb]
  \centering
  \includegraphics[width=\linewidth]{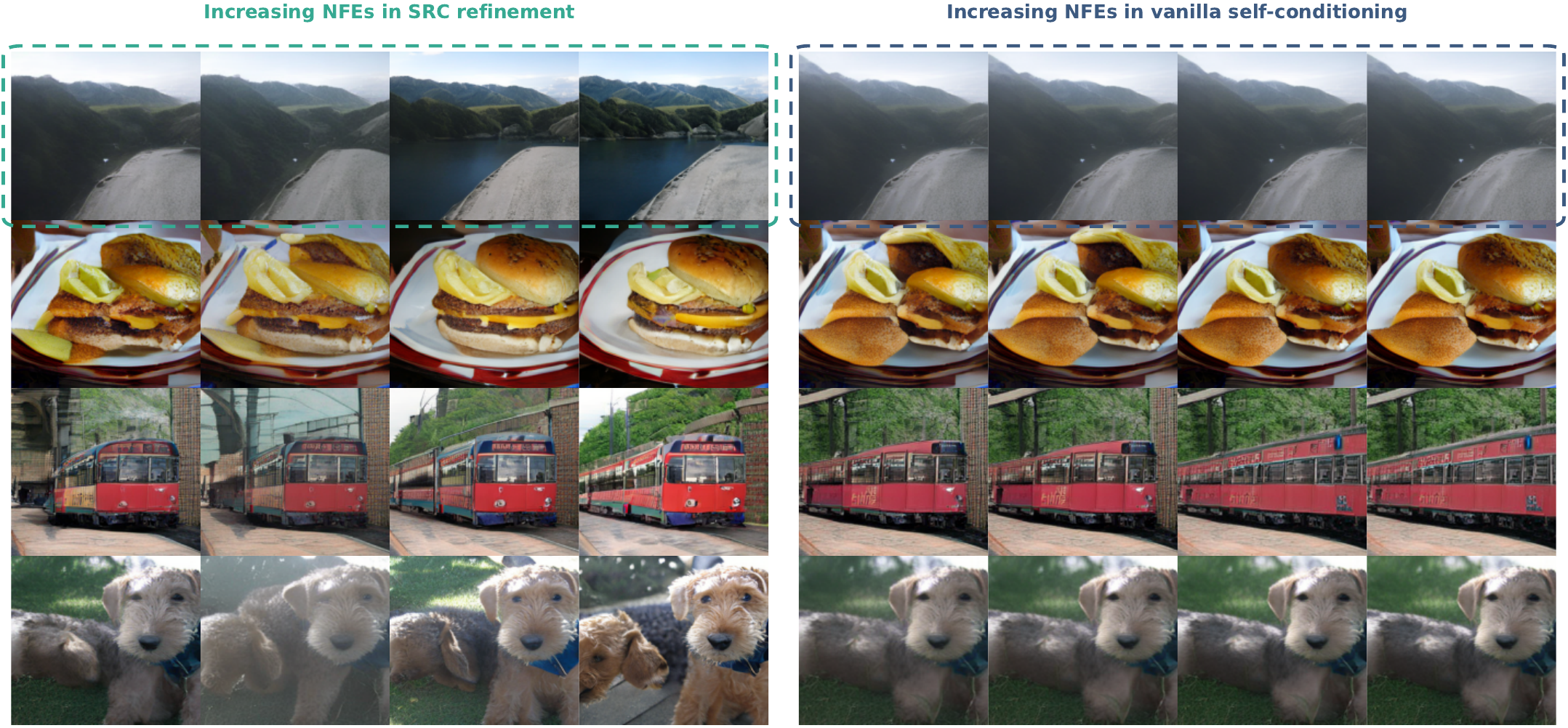}
  \caption{Matched-seed comparison of \src{} refinement and vanilla
  self-conditioning refinement. Left: \src{} at NFE
  $\{50,100,200,550\}$. Right: vanilla self-conditioning at the same NFEs
  (reference, $R{=}1$, $R{=}3$, $R{=}10$). The right-hand family is the
  baseline that plateaus in Fig.~\ref{fig:inplace-plateau}.}
  \label{fig:src-vsc}
\end{figure*}

Figure~\ref{fig:src-vsc} compares \src{} refinement with vanilla
self-conditioning refinement on the same seeds and classes. The
self-conditioning side is the $R$-scaling baseline whose FID plateaus in
Fig.~\ref{fig:inplace-plateau}. Columns are matched in NFE. On the left,
\src{} uses $K{\times}R$ settings at NFE 50, 100, 200, and 550. On the
right, vanilla self-conditioning uses its reference step at NFE 50, then
$R{=}1$, $R{=}3$, and $R{=}10$, which land at NFE 100, 200, and 550.
Extra rounds of in-place refinement leave the sample close to where it
started. \src{} keeps composition and layout while improving readability and coherence of the object in the image.

\end{document}